%% file: acl.tex
\pdfoutput=1
\documentclass[11pt]{article}

\usepackage{acl}

\usepackage{times}
\usepackage{latexsym}
\usepackage{array}
\usepackage{booktabs}
\usepackage{multirow}
\usepackage{siunitx}
\input{init}

\usepackage[T1]{fontenc}
\usepackage[utf8]{inputenc}

\usepackage{microtype}

\usepackage{graphicx}

\title{Identifying and Measuring Token-Level Sentiment Bias in Pre-trained Language Models with Prompts}

\author{Apoorv Garg$^*$, \  Deval Srivastava\thanks{$^*$ Apoorv Garg and Deval Srivastava contributed equally to this work.}, \ Zhiyang Xu, \ Lifu Huang
\\
  Computer Science Department, \ Virginia Tech
 \\
  {\tt \{apoorvgarg,devalsrivastava,zhiyangx,lifuh\}@vt.edu},
  }
  
\begin{document}
\maketitle

\input{0abstract}
\input{1introduction}
\input{2relatedwork}

\input{3approach}

\input{4experiment}

\input{5discussion}
\input{6conclusion}

% \section*{Acknowledgements}
% This document has been adapted

\bibliography{anthology,custom}
\bibliographystyle{acl_natbib}

\appendix

\section{Appendix}
\label{sec:appendix}
\subsection{Implementation Details}
\label{ssec:Implementation}
We use the pre-trained RoBERTa-Base and RoBERTa-Large models from Huggingface. We use the Adam optimizer with a learning rate of 1e-5 and batch size 2 to train our models. Each epoch takes about 30 mins and we run the experiment on one Tesla P40.
% This is an appendix.

\subsection{Impact of Prompt Templates}
\label{ssec:prompt_templates}
% Prompt plays a important role in our work, we classify the sentiment of a text segment using a prompt template. In our experiments we try prompts like ``\texttt{the review was [MASK]}'',``\texttt{I [MASK] it.}'' et cetera but found them to be worse. we use ``\texttt{It was [MASK].}'' which allows the model to understand the most sentiment polarity. We use the prompt labels ``\textit{great}'' and ``\textit{terrible}''. Use of this prompt template and and these prompt labels are also backed by \cite{DBLP:conf/acl/GaoFC20} where they researched prompt templates and found similar results.
% We experimented with ``\textit{good}'', ``\textit{bad}'' and  ``\textit{like}'', ``\textit{dislike}'' and found them to be worse. 
As the prompt plays an important role in our work, we experiment with different types of templates, such as  ``\texttt{the review was [MASK]}'' and ``\texttt{I [MASK] it.}'' but we find that adding more context affects the models' ability to identify bias. The context dominates the prediction. Thus, we decide to use the simple template ``\texttt{It was [MASK].}'' in all of our experiments.
% 
% As the prompt plays an important role in our work, we experiment with different types of templates, such as  ``\texttt{the review was [MASK]}'' and ``\texttt{I [MASK] it.}'' but we find that adding more context affects the models' ability to identify bias. The context dominates the prediction. Thus, we decide to use the simple template ``\texttt{It was [MASK].}'' in all of our experiments.

\subsection{Experiment Results}
\label{appen_exp}

\begin{table*}[ht!]
\begin{center}
\caption{This table shows the top-10 most biased words identified by SAT on various PLMs. The \% column shows the percentage of identified biased words in all the neutral words (400).}
\resizebox{\textwidth}{!}{%
\begin{tabular}{{p{2.0cm}p{1.9cm}p{4.0cm}p{0.6cm}p{6cm}p{0.5cm}}}
\hline
Model &Threshold &Positive Words &\% &Negative Words &\%\\
\hline
RoBERTa-Base & 0.5*Standard Deviation & \makecell{modular,shone,Tours}&0.75& \makecell{obvious,deadly,vanilla,\\cheese,speculation,systematic,\\conjecture,sleepy,economics,vodka}&17.0\\
RoBERTa-Base & 1*Standard Deviation & \makecell{shone}&0.25 & \makecell{cheese,speculation,turkey,\\skeletal,outright,bacterial,\\carrots,tires,dialog,attitudes}&3.75\\
RoBERTa-Base & 1.5*Standard Deviation & \makecell{-}&0.0 & \makecell{cheese,speculation,carrots}&0.75\\
Finetuned RoBERTa-Base & 0.5*Standard Deviation & \makecell{PEOPLE,sovereignty,\\ embodiment,predominant,\\Indeed,incorporate,touch}&1.75&\makecell{vanilla,skeletal,speculation,\\implicit,cheese,sleepy,overweight,\\pitched,judgement,conjecture}&36.25\\
Finetuned RoBERTa-Base & 1*Standard Deviation & \makecell{sovereignty}&0.25& \makecell{vanilla,skeletal,speculation,\\cheese,sleepy,overweight,conjecture,\\rural,Possible,turkey}&24.0\\
Finetuned RoBERTa-Base  & 1.5*Standard Deviation & \makecell{-}&0.0& \makecell{vanilla,skeletal,speculation,\\cheese,sleepy,overweight,conjecture,\\rural,Possible,turkey}&16.25\\
RoBERTa-Large & 0.5*Standard Deviation & \makecell{shone}&0.25 & \makecell{deadly,screaming,cheese,\\speculation,overtime,overweight,\\implicit,systematic,vodka,conjecture}&12.5\\
RoBERTa-Large & 1*Standard Deviation & \makecell{shone}&0.25 & \makecell{cheese,speculation,conjecture,\\bacterial,tires,signals,\\ander,Minor}&2.0\\
RoBERTa-Large & 1.5*Standard Deviation & \makecell{-}&0.0& \makecell{bacterial}&0.25\\
Finetuned RoBERTa-Large & 0.5*Standard Deviation & \makecell{precious,shone,servings,\\embodiment,Indeed}&1.25& \makecell{familiar,implicit,screaming,\\overweight,cheese,sleepy,speculation,\\bacterial,turkey,conjecture}&29.5\\
Finetuned RoBERTa-Large & 1*Standard Deviation & \makecell{shone}&0.25 & \makecell{implicit,overweight,sleepy,speculation,\\bacterial,conjecture,patriarchal,\\Possible,vodka,glare}&16.5\\
Finetuned RoBERTa-Large & 1.5*Standard Deviation & \makecell{-}&0.0& \makecell{implicit,overweight,sleepy,\\speculation,bacterial,turkey,conjecture,\\patriarchal,Possible,vodka}&9.0\\
\hline
\end{tabular}
\label{table:assoc_test_results}
}
\end{center}
\end{table*}

% \begin{table*}[ht]
% \centering
% \caption{Top 10 words for each model for sentiment shift test}

% \begin{tabular}{l r r r r}
% \hline
% Model & Positive Words & \% & Negative Words & \% \\
% \hline
% RoBERTa-large Finetuned &familiar , shone , comedy , vanilla , renewable , exercised , consistency , insights , chocolate , convertible & 44.5 &  judgement, patriarchal, assumption, Minor, overweight, conjecture,
%       distance, bacterial, appeal, stall & 44.8 \\
% RoBERta-large & renewable , exercised , sovereignty , precious , Jordanian , intrigue , modular , comedy , Destiny , Episcopal & 63.2 & deadly , judgement , appeal , disposition , skeletal , patriarchal , corrective , December , systematic , convict & 23.17\\
% RoBERTa-base & shone , dominant , clout , intrigue , globalization , uncover , Saint , Circle , Beans , exercised & 24.2 & deadly , judgement , screaming , plight , appeal , bacterial , glare , obligations , obligation , overweight & 42.5\\

% Roberta-base Finetuned & shone , extensive , Saint , Indeed , systematic , Awareness , concerted , insights , dominant , renewable & 44.5 &  deadly , judgement , overweight , speculation , glare , appeal , patriarchal , tobacco , adversity , notion & 17.6\\

% \hline
% \end{tabular}
% \label{table:sentiment_shift_results}
% \end{table*}

\begin{table*}[ht!]
\centering
\caption{This table shows the top-10 most biased words identified by SST on various PLMs, with k=5. The \% column shows the percentage of identified biased words in all the neutral words (400).}
\resizebox{\textwidth}{!}{%
\begin{tabular}{{p{2.5cm}p{0.5cm}p{5.5cm}p{0.5cm}p{5.5cm}p{0.5cm}}}
\hline
Model & K & Positive Words & \% & Negative Words & \% \\
\hline

RoBERTa-base & 5 & shone , dominant , clout , intrigue , globalization , uncover , Saint , Circle , Beans , exercised & 57 & deadly , judgement , screaming , plight , appeal , bacterial , glare , obligations , obligation , overweight & 31\\

RoBERTa-base & 10 & anyways , utilizes , shone , dominant , attaches , intrigue , clout , uncover , reflecting , globalization & 58 & deadly , judgement , screaming , undergoing , glare , appeal , opinions , speculation , plight , referee & 34 \\ 

RoBERTa-base & 15 & anyways , clout , dominant , intrigue , utilizes , attaches , shone , uncover , incorporate , reflecting & 59.2 & deadly , judgement , screaming , speculation , undergoing , glare , counselling , referee , subsequently , Possible & 34.5 \\ 

Finetuned RoBERTa-base  & 5 & shone , extensive , Saint , Indeed , systematic , Awareness , concerted , insights , dominant , renewable & 44.5 &  deadly , judgement , overweight , speculation , glare , appeal , patriarchal , tobacco , adversity , notion & 17.6\\

Finetuned RoBERTa-base  & 10 & shone , extensive , Saint , quite , dominant , concerted , renewable , Awareness , insights , Indeed & 44 &  deadly , judgement , overweight , speculation , appeal , glare , tobacco , screaming , speculate , patriarchal & 27.5\\

Finetuned RoBERTa-base  & 15 & extensive , shone , Saint , quite , dominant , insights , incorporate , participants , concerted , entirely & 45 &  judgement , deadly , speculation , overweight , appeal , notified , speculate , counselling , glare , screaming & 30\\

RoBERta-large & 5 & renewable , exercised , sovereignty , precious , Jordanian , intrigue , modular , comedy , Destiny , Episcopal & 63.2 & deadly , judgement , appeal , disposition , skeletal , patriarchal , corrective , December , systematic , convict & 23.17\\

RoBERta-large & 10 & Destiny , precious , renewable , Rapid , sovereignty , Jordanian , Awareness , intrigue , incorporate , Episcopal & 60.2 &  deadly , judgement , disposition , appeal , patriarchal , screaming , Pricing , skeletal , plight , glare & 23\\

RoBERta-large & 15 &Awareness , Destiny , incorporate , Episcopal , precious , renewable , intrigue , Jordanian , sovereignty , olive & 59.5 & deadly , judgement , disposition , appeal , screaming , patriarchal , glare , counselling , skeletal , Confederate & 22 \\

Finetuned RoBERTa-large  & 5 & familiar , shone , comedy , vanilla , renewable , exercised , consistency , insights , chocolate , convertible & 44.5 &  judgement, patriarchal, assumption, Minor, overweight, conjecture,
      distance, bacterial, appeal, stall & 44.8 \\

Finetuned RoBERTa-large & 10 & familiar , insights , vanilla , consistency , shone , olive , chocolate , exercised , correctness , silver & 48.1 & judgement , deadly , disposition , patriarchal , assumption , distance , appeal , conjecture , overweight , impacts & 41.8 \\

Finetuned RoBERTa-large & 15 & familiar , insights , vanilla , consistency , shone , olive , correctness , silver , incorporate , extensive & 49.8 &
judgement , deadly , disposition , assumption , conjecture , distance , appeal , patriarchal , glare , overweight & 39\\

\hline
\end{tabular}
\label{table:sentiment_shift_results}
}
\end{table*}

\begin{table}[ht]
\centering
\caption{Neutral words obtained using SST on the finetuned Roberta-Base}
\begin{tabular}[t]{lc}
\hline
Word & Score\\
\hline
Whites & -0.0\\quarter & -0.01\\System & -0.04\\Posts & -0.05\\bucks & -0.05\\
Religious & 0.0\\downright & 0.01\\outcome & 0.01\\Count & 0.02\\events & 0.02\\
\hline
\end{tabular}
\label{table:toprankwords1}
\end{table}

% \newline
\begin{table}[ht]
\centering
\caption{Top 10 most positive-biased words using SST on finetuned Roberta-Base}
\begin{tabular}[t]{lc}
\hline
Word & Score\\
\hline
shone & 20.62\\extensive & 15.17\\Saint & 13.24\\Indeed & 9.67\\Awareness & 9.39\\systematic & 8.92\\concerted & 8.81\\dominant & 8.63\\insights & 8.62\\renewable & 8.18\\
\hline
\end{tabular}
\label{table:toprankwords2}

\end{table}

\begin{table}[ht]
\centering
\caption{Top 10 most negative-biased words using SST on finetuned Roberta-Base}
\begin{tabular}[t]{lc}
\hline
Word & Score\\
\hline
deadly & -24.48\\judgement & -19.13\\overweight & -14.57\\speculation & -13.97\\glare & -11.59\\appeal & -9.94\\patriarchal & -8.4\\tobacco & -7.92\\screaming & -6.66\\replacing & -6.31\\

\hline
\end{tabular}
\label{table:toprankwords3}

\end{table}

\begin{table}[ht]
\centering
\caption{Top 10 most positive-biased words using SST on finetuned Roberta-Base}
\begin{tabular}[t]{lc}
\hline
Word & Score\\
\hline
shone & 15.52\\dominant & 12.67\\clout & 12.49\\intrigue & 11.37\\globalization & 10.59\\uncover & 10.49\\Circle & 9.71\\Saint & 9.26\\Beans & 9.03\\reflecting & 8.73\\
\hline
\end{tabular}
\label{table:toprankwords4}

\end{table}

\begin{table}[ht]
\centering
\caption{Top 10 most negative-biased words using SST on finetuned Roberta-Base}
\begin{tabular}[t]{lc}
\hline
Word & Score\\
\hline
deadly & -36.74\\judgement & -30.35\\screaming & -18.26\\plight & -16.29\\glare & -14.28\\bacterial & -14.24\\appeal & -10.89\\obligations & -10.87\\referee & -10.46\\obligation & -10.14\\

\hline
\end{tabular}
\label{table:toprankwords5}

\end{table}

\begin{table}[ht]
\centering
\caption{Neutral words obtained using SST on the pre-trained Roberta-Base}
\begin{tabular}[t]{lc}
\hline
Word & Score\\
\hline
puppy & -0.04\\silver & -0.07\\expectation & -0.09\\Productions & -0.1\\bucket & -0.12\\
Hindu & 0.02\\Fiscal & 0.03\\stances & 0.05\\notion & 0.06\\supplies & 0.1\\
\hline
\end{tabular}
\label{table:toprankwords6}

\end{table}

\end{document}

%% file: init.tex
\usepackage{graphicx}               % Include graph
\usepackage{tabularx}               % Better table formatting
\usepackage{booktabs}
\usepackage{amsmath}
\usepackage{stmaryrd}
\usepackage{xcolor}
\usepackage{soul}

\newcolumntype{C}{>{\centering\arraybackslash}X}
\usepackage{multirow}               % Multi-row in tables
\usepackage{diagbox}                % Diagonal line in tables
\usepackage{hhline}                 % Draw double line in tables
\usepackage{color}                  % Text and background color
\usepackage{amsmath}                % Formula
\usepackage{amssymb}                % Formula
\usepackage{mathtools}              % Formula
\usepackage{enumitem}               % Better itemize environment

\usepackage{subfigure}
\usepackage{booktabs}
\usepackage{extarrows}
\usepackage{makecell}

\definecolor{RoseQuartzBg}{HTML}{F7CAC9}
\definecolor{RoseQuartz}{HTML}{F5A798}
\definecolor{Serenity}{HTML}{92A8D1}
\definecolor{OrangeRed}{rgb}{1.0, 0.27, 0.0}
\definecolor{Red}{rgb}{1.0, 0.0, 0.0}
\definecolor{Turquoise}{HTML}{0F4C81}
\usepackage{xparse}
\NewDocumentCommand{\lifu}{ mO{} }{\textcolor{OrangeRed}{\textsuperscript{\textit{Lifu}}\textsf{\textbf{\small[#1]}}}}
\NewDocumentCommand{\zhiyang}{ mO{} }{\textcolor{blue}{\textsuperscript{\textit{zhiyang}}\textsf{\textbf{\small[#1]}}}}
\NewDocumentCommand{\deval}{ mO{} }{\textcolor{Red}{\textsuperscript{\textit{Deval}}\textsf{\textbf{\small[#1]}}}}
\NewDocumentCommand{\apoorv}{ mO{} }{\textcolor{Red}{\textsuperscript{\textit{Apoorv}}\textsf{\textbf{\small[#1]}}}}

%% file: 0abstract.tex
\begin{abstract}
% In the recent years there has been a lot of research on demonstrating, quantifying and mitigating various biases  in language model. In this paper we research on a different bias known as Sentiment Bias. We propose 2 methods to demonstrate that some of the currently used language models have this bias and also find certain words that directly have this bias. We then design an experiment to show the impact of sentiment bias on a commonly used downstream task of sentiment classification, Using this experiment we quantify the strength of sentiment bias shown by a neutral word. We also conduct various tests to demonstrate that the models considered for the above experiments are capable of understanding positive and negative sentiment polarity of a particular sentiment segment entry.
% Our approach uses Pretrained Language Models with MLM heads and leverages the entire inferred logit space for prompts rather than the most probable logit to associate bias with the PLM.

Due to the superior performance, large-scale pre-trained language models (PLMs) have been widely adopted in many aspects of human society. However, we still lack effective tools to understand the potential bias embedded in the black-box models. Recent advances in prompt tuning show the possibility to explore the internal mechanism of the PLMs. In this work, we propose two token-level sentiment tests: Sentiment Association Test (SAT) and Sentiment Shift Test (SST) which utilize the prompt as a probe to detect the latent bias in the PLMs. Our experiments on the collection of sentiment datasets show that both SAT and SST can identify sentiment bias in PLMs and SST is able to quantify the bias. The results also suggest that fine-tuning can possibly augment the existing bias in PLMs.
%\lifu{assumption: the reviews can fully represent the positive and negative sentiment}

%\lifu{Question0: what are the best strategies to represent the positive and negative sentiment?}

%\lifu{Question1: are there any sentiment bias in the pre-trained language models/ fine-tuned language models?}

%\lifu{Question2: what is the difference among the sentiment bias of different pre-trained language models?}

%\lifu{Question3: is there a correlation between sentiment bias with other types of bias? - collect a set of words related to other bias, gender bias, political bias}

%\lifu{Question4: how to mitigate the sentiment bias?}
\end{abstract}

%% file: 1introduction.tex
\section{Introduction}

Large-scale pre-trained language models (PLMs), such as BERT~\cite{DBLP:conf/naacl/DevlinCLT19}, RoBERTa~\cite{DBLP:journals/corr/abs-1907-11692}, GPT~\cite{radford2018improving,radford2019language,DBLP:conf/nips/BrownMRSKDNSSAA20} and T5~\cite{DBLP:journals/jmlr/RaffelSRLNMZLL20}, have shown competitive performance in many downstream applications in natural language processing. The key to the success of PLMs lies in the unsupervised pre-training on massive unlabeled corpus as well as a large number of parameters in the neural models. While these PLMs have been deployed to a wide variety of products and services such as search engines and chatbots, investigating the fairness of these PLMs has become a growing urgent research agenda.

Recent studies have shown that there are various stereotypical biases related to social factors such as gender~\cite{DBLP:journals/cogcom/BhardwajMP21}, race~\cite{DBLP:journals/corr/abs-2006-11316}, religion~\cite{DBLP:conf/acl/NadeemBR20}, age~\cite{DBLP:conf/emnlp/NangiaVBB20}, ethnicity~\cite{DBLP:conf/emnlp/GroenwoldOPHLMW20}, political identity~\cite{DBLP:journals/corr/abs-2009-06807}, disability~\cite{DBLP:journals/corr/abs-2005-00813}, name~\cite{DBLP:journals/corr/abs-2004-03012} and many more, that are inherited by these PLMs. However, sentiment bias, which characterizes the bias of words towards a particular sentiment polarity, such as \textit{positive} or \textit{negative}, has not been well studied, yet it has significant impact to many applications, such as market analysis or chat bots, where subtle bias may lead to distinct understanding or conclusions.~\newcite{DBLP:conf/emnlp/HuangZJSWRMYK20} investigated the sentiment bias in texts generated by language models like GPT while overlooking the fact that each individual word may also have sentiment bias in the PLMs. 
%Even though less studied, it has significant impact to many applications, such as .. in sentiment classification applications on social media, chat bots and generation tasks.

% in their work they define a conditioning text and generate text based on that, they show that by varying some sensitive attributes in the conditioning texthe sentiment polarity of the generated text can change. 

In this work, we focus on identifying and measuring the sentiment bias of individual words in pre-trained language models. Instead of investigating all the words in the vocabulary, we only select a list of words with confident sentiment polarities from available sentiment lexicons constructed by humans, and design two novel approaches to identify their sentiment bias based on language model prompting: (1) Sentiment Association Test (SAT), where the bias of each word is identified by detecting its association with various positive or negative reviews; (2) Sentiment Shift Test (SST), where the bias of each word is identified by predicting the sentiment polarity shift after appending it multiple times to various sentiment-oriented reviews. Based on these two approaches, we observe that 39.25\% out of 400 words considered neutral in the lexicon show a sentiment bias in commonly used PLMs. In addition, by extending the Sentiment Shift Test, we further design a new metric to measure the strength of the sentiment bias for each word. Our contributions are summarized as follows:

\begin{itemize}%[noitemsep,nolistsep,wide]
\item We design two novel sentiment test approaches, SAT and SST, to investigate the token-level sentiment bias from the PLMs, and demonstrate that 39.25\% out of 400 neutral words show a sentiment bias in various PLMs. 

\item We also design a new metric to quantify the sentiment bias of each word by extending our Sentiment Shift Test.
\end{itemize}

%% file: 2relatedwork.tex
\section{Related Work}

% \lifu{please try to summarize the related work for each of the following categories. You can refer to related papers, e.g., \url{https://arxiv.org/pdf/1911.03064.pdf}, \url{https://mdpi-res.com/d_attachment/applsci/applsci-11-03184/article_deploy/applsci-11-03184-v2.pdf}}
% \lifu{can you add more references for each of the following paragraph, especially the first two. For each approach, just use one sentence to summarize the key idea of it. And group them into categories based on their research topics if possible.}

% \paragraph{Stereotypical Bias in Natural Language Processing}
% When we train these large language models on huge text corpuses, these models may pick up social biases like stereotypes.\newcite{DBLP:conf/acl/NadeemBR20} define bias based on gender, profession, race and religion and design a formulae to quantify the stereotypical bias along with model meaningfulness of PLMs for sentence level and discourse level reasoning. In a related work authors of CrowS-pairs \newcite{DBLP:conf/emnlp/NangiaVBB20}, design another dataset to detect social biases in language models. In comparison to Steroset they define a metric on how likely is for the stereotype/anti-stereotype to generate the rest of the sentence.\newcite{DBLP:journals/corr/abs-2005-00813} use Google Cloud Sentiment model to  demonstrate more negative bias in top-k words predicted by BERT when prompted with disability tokens.
\paragraph{Stereotypical Bias in Natural Language Processing}
\newcite{DBLP:conf/acl/NadeemBR20} define bias based on gender, profession, race, and religion, and design formulae to quantify the stereotypical bias along with model meaningfulness of PLMs for sentence-level and discourse-level reasoning.
\newcite{DBLP:conf/emnlp/NangiaVBB20} define a metric on how likely is the stereotype/anti-stereotype to generate the rest of the sentence. \newcite{DBLP:journals/corr/abs-2005-00813} use the Google Cloud Sentiment model to demonstrate more negative bias in top-k words predicted by BERT when prompted with disability tokens.
%  They argue that when we predict the mask as in stereoset we run the risk of predicting the token more occuring in the data.
\paragraph{Bias in natural language embeddings}

% In the past few years there has been a lot of research in the NLP domain to find bias within word embeddings, Approaches have been designed using analogy tests \newcite{DBLP:conf/nips/BolukbasiCZSK16}, where the authors demonstrate that word2vec embeddings reflect gender bias, they show that female word are closer to gender stereotypical words. authors \newcite{DBLP:journals/corr/IslamBN16} proposed WEAT ( Word embedding association test ) to demonstrate how names can be associated with entities. They show that European names were more associated with pleasant words compared to African names associated with unpleasant words and that female names are  associated with familial words rather than occupations, indicating bias. \newcite{DBLP:journals/corr/abs-1904-03310} demonstrate that contextualized word embeddings have bias. The authors demonstrate that ELMo has been trained on a skewed corpus, furthermore they show that the contextualized embeddings are biased. Then they show how the bias propagates to a downstream like co-reference resolution on the WinoBias dataset.
Many studies explore bias within word embeddings. \newcite{DBLP:conf/nips/BolukbasiCZSK16} utilize analogy tests and demonstrate that word2vec embeddings reflect gender bias by showing that female names are associated with familial words rather than occupations. \newcite{DBLP:journals/corr/IslamBN16} proposed WEAT ( Word embedding association test ) to show how names can be associated with entities. \newcite{DBLP:journals/corr/abs-1904-03310} prove that contextualized word embeddings have bias and show how bias propagates to downstream tasks.

\paragraph{Identification and Measurement of Sentiment Bias}

% \newcite{DBLP:conf/emnlp/HuangZJSWRMYK20}, propose detection of sentiment bias by varying some sensitive attributes and measure the sentiment polarity of the generated text using GPT-2. Similarly, \newcite{DBLP:conf/emnlp/GroenwoldOPHLMW20} determine sentiment bias for ethnicity by comparing the sentiment of text generated by GPT-2 and find more negative sentiment generated for African American Vernacular English text as compared to Standard American English text.

\newcite{DBLP:conf/emnlp/HuangZJSWRMYK20} propose detection of sentiment bias by varying some sensitive attributes and measuring the sentiment polarity of the generated text using GPT-2.
\newcite{DBLP:conf/emnlp/GroenwoldOPHLMW20} determine sentiment bias for ethnicity by comparing the sentiment of text generated by GPT-2 and find more negative sentiment generated for African American Vernacular English text as compared to Standard American English text. Compared to the above studies, our work investigates and measures the sentiment bias at token level from PLMs.

%% file: 3approach.tex
\section{Approach}

\subsection{Dataset Construction}
\label{sec:dataset}
% Our goal is to investigate and measure the sentiment bias of each individual word in PLMs. Considering that many words may indicate distinct sentiment polarities in different context, we first build a highly confident sentiment lexicon where each word is annotated as \textit{positive}, \textit{negative} or \textit{neutral}. Specifically, we combine multiple sentiment lexicons that are annotated by human, including the VADER lexicon~\cite{DBLP:conf/icwsm/HuttoG14} where all the words are annotated with sentiment scores from -4 to +4 (-4 being strongly negative and +4 being strongly positive) and the MPQA opinion corpus~\cite{DBLP:conf/naacl/DengW15}.\lifu{are the last two lexicons also including sentiment scores? If not, how did you select the confident words?}.\deval{ Based on the sentiment scores, we finally selected top ranked 400 positive, negative words from the VADER lexicon and neural words from the MPQA opinion corpus} \lifu{update this number if necessary} , respectively and investigate the sentiment bias only on neutral words.We estimate sentiment polarity based on positive and negative words.
Our goal is to investigate and measure the sentiment bias of each word in PLMs. Considering that many words may indicate distinct sentiment polarities in different context, we first build a highly confident sentiment lexicon where each word is annotated as \textit{positive}, \textit{negative} or \textit{neutral}. Specifically, we draw strongly positive and negative tokens from the VADER lexicon~\cite{DBLP:conf/icwsm/HuttoG14} where all the words are annotated with sentiment scores from -4 to +4 (-4 being strongly negative and +4 being strongly positive). We draw the neutral words from MPQA opinion corpus~\cite{DBLP:conf/naacl/DengW15}.%\lifu{are the last two lexicons also including sentiment scores? If not, how did you select the confident words? \deval{The other lexicon has a neutral section}}.
% \deval{ Based on the sentiment scores, we finally selected top ranked 400 positive, negative words from the VADER lexicon and neural words from the MPQA opinion corpus} \lifu{update this number if necessary} , 
We investigate the sentiment bias only on the neutral words, and use the positive and negative words to verify our approaches.  

% Later, we will only investigate the sentiment bias of these words.

%To achieve this goal, we first construct a highly confident sentiment lexicon where each word is annotated as \textit{positive}, \textit{negative} or \textit{neutral}. Specifically, we combine the VADER lexicon~\cite{VADERlexicon}\lifu{add reference or a link} where all the words are annotated with sentiment scores in the range of -4 to +4 (-4 being strongly negative and +4 being strongly positive), MPQA opinion corpus~\cite{mpqa2015corpus}\lifu{add reference or a link} and the Opinion lexicon~\cite{liu2004opinioncorpus}\lifu{add reference or a link}. Based on the sentiment scores, we finally selected 400 positive words, 400 negative words, and 400 neutral words. Later, we will just investigate the sentiment bias of these words.

% In addition, the sentiment bias of each word is being identified conditional on the sentiment polarity which are learned from multiple popularly used sentiment classification datasets,

The sentiment lexicon contains a golden sentiment label of each word. Thus, to detect the sentiment bias, we need to compare the sentiment polarity of each word in PLMs with its golden sentiment label. To predict the sentiment polarity of each word in PLMs, we will leverage a set of sentiment-oriented reviews collected from IMDB~\cite{DBLP:conf/acl/MaasDPHNP11}, Amazon Reviews~\cite{DBLP:conf/www/HeM16}, YELP~\cite{DBLP:journals/corr/Asghar16}, and SST-2~\cite{DBLP:conf/emnlp/SocherPWCMNP13}. Each review is annotated as positive or negative. We collect 2000 positive reviews and 2000 negative ones. As the reviews span over diverse domains, including movies, food, and products, they can well represent each sentiment polarity.
% \lifu{update this numbeer}
% \lifu{update this number} 
%To identify the sentiment bias, we need to analyze the association of each word from the sentiment lexicon with each sentiment polarity based on the encoding of pre-trained language models. To represent each sentiment polarity, we will further leverage the available sentiment classification datasets, including IMDB~\cite{IMDBreviews}, Amazon Reviews~\cite{AmazonReviews}, YELP~\cite{} and SST-2~\cite{SST2dataset}, where reviews are annotated as positive or negative and cover diverse domains, such as movies, food reviews, product reviews, so that they can well represent each sentiment polarity. Finally, we collected 2000 reviews for positive polarity, and 2000 for negative.

\subsection{Sentiment Bias Identification}

\begin{figure}
\centering
\includegraphics[scale=0.85]{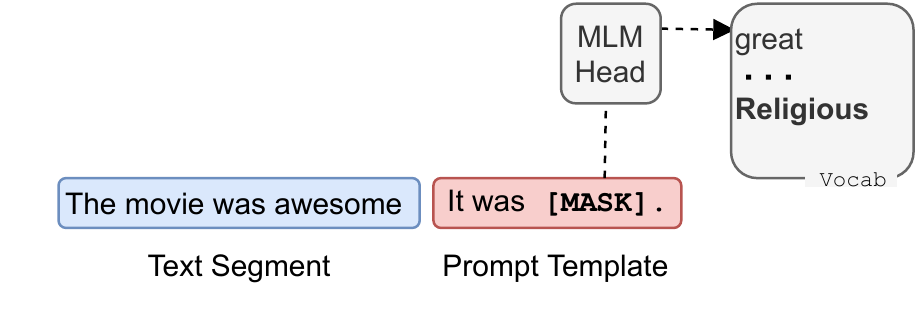}
\caption{Overview of the prompting approach for SAT.}
\label{fig:1}
\vspace{-1.4em}
\end{figure}

\paragraph{Sentiment Association Test} Inspired by the Word Embedding Association Test~\cite{DBLP:journals/corr/IslamBN16}, we first design a new Sentiment Association Test approach to predict the sentiment polarity of each word in PLMs based on their associations.
Our approach is based on the assumption that if a word consistently shows a stronger association to the diverse set of positive (or negative) reviews, it should have a positive (or negative) sentiment polarity. Based on this assumption, we design a language model prompting approach to estimate the association of each word with a review. As Figure~\ref{fig:1} shows, given each positive review $p_{i}$ , we concatenate it with a template-based prompt, ``\texttt{It was [MASK]}'', and feed the whole sequence to a language model encoder. Based on the contextual representation of ``\texttt{[MASK]}'', we predict a probability for each word in the sentiment lexicon as $s_{ij}^{p}$ where $j$ is the index of the word in the lexicon. Similarly, for each negative review $n_{k}$, we apply the same prompt and use the same approach to predict a probability for each word in the lexicon as $s_{kj}^{n}$. For each word indexed with $j$, we determine its sentiment polarity by comparing $mean_i(s_{ij}^{p})$  with $mean_k(s_{kj}^{n}) + m*std_k((s_{kj}^{n}))$ and $mean_k(s_{kj}^{n})$  with $mean_i(s_{ij}^{p}) + m*std_i((s_{ij}^{p}))$, which denotes the association to positive and negative sentiment polarity, respectively. $std$ denotes the standard deviation, which in this case serves as a dynamic unit to measure the distance of the means between the positive and negative probability mass functions (PMFs) and $m$ shows the strength of the sentiment polarity. With a fixed unit, or no dynamic unit, we can only identify the strongly biased words.

\begin{figure}
\centering
\includegraphics[scale=0.7]{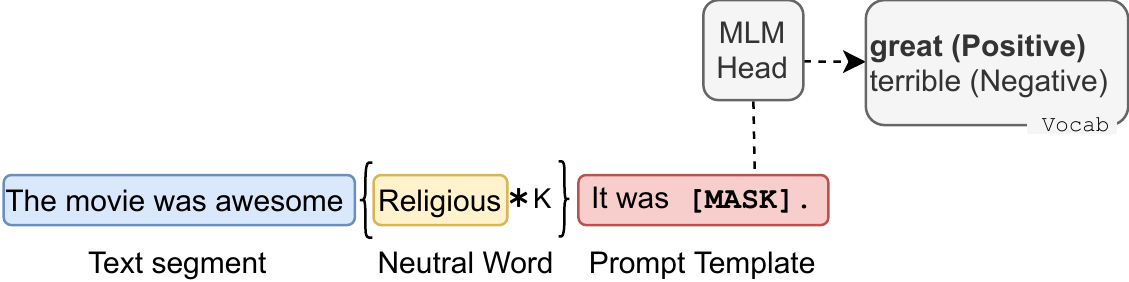}
\caption{Overview of the prompting approach for SST.}
\label{fig:2}
\vspace{-1.3em}
\end{figure}

\paragraph{Sentiment Shift Test} Another intuitive approach to predict the sentiment bias of each word in PLMs is based on the assumption that if a word is negative in PLMs and appended multiple times to a positive review, it's likely that the sentiment of this new sequence might be shifted to neutral or even negative. Based on this assumption, we further design a new Sentiment Shift Test approach to predict the sentiment bias of each word in PLMs. As Figure~\ref{fig:2} shows, given a review, we first apply language model prompting to concatenate the review with a prompt ``\texttt{It was [MASK]}'', and predict a sentiment label by comparing the probability of ``\textit{great}'' and ``\textit{terrible}'' based on the contextual representation of ``\texttt{[MASK]}''. Then, for each word in the lexicon, we append it $K$ times to the review, and use the same language model prompting approach to predict a sentiment label. We will predict the sentiment bias of each word in PLMs by analyzing the number of sentiment shifts for all positive or negative reviews, i.e., if a word is appended to positive reviews and reduces the accuracy of the model on reviews, this word will have a negative bias.

\subsection{Sentiment Bias Quantification}
Based on the Sentiment Shift Test, we further design a new metric to quantify the strength of the sentiment bias of each word. Our motivation is that, the less times that a word is appended and the more sentiment labels are shifted after appending it to the reviews, the stronger that the bias will be. Based on this motivation, we design the following metric:
$q = \frac{1}{n}\sum_{K}\frac{(Neg\_Diff - Pos\_Diff)}{K^2}$
where, $Neg\_Diff = A - A^{'}$ is defined as the change of the sentiment classification accuracy on the negative sentiment set after appending a neutral word K times. $A$ and $A^{'}$ denote the accuracy before and after appending the word, respectively.
%e.g. the model predicted 1400 to be negative out of total 2000 negative texts originally, when we add a neutral word K times the model predicted 1200 to be negative, then the $Negative Diff$ would be 200.  
Similarly, $Pos\_Diff$ is defined as the change of the accuracy on the positive sentiment set. 
If a neutral word reduces the negative accuracy, $Neg\_Diff$ will be positive and $Pos\_Diff$ is more likely to be negative, providing us with a high overall positive value implying a positive sentiment bias.

%% file: 4experiment.tex
\section{Experimental Results and Discussion}

\subsection{Experimental Setup}

We select 400 words for each of positive, negative, and neutral categories from the sentiment lexicon and perform SAT and SST on the 2,000 positive and 2,000 negative reviews. The experiments are mainly on RoBERTa models as they show a significantly better understanding of sentiment presented in the text than BERT models. We analyze the word-level sentiment bias in both the pre-trained language model and prompt\footnote{We study the impact of different prompt templates and label words in Appendix \ref{ssec:prompt_templates}}-based fine-tuning model. The prompt-based model follows the training framework from \cite{DBLP:conf/acl/GaoFC20} which utilizes a set of training instances as demonstrations to help the model make predictions.

\subsection{Does the Probability Predicted by the Language Model Indicate the Sentiment Polarity?}

\begin{figure}[h]
\centering
  \includegraphics[width=0.48\textwidth]{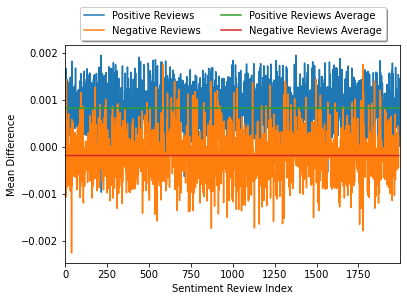}
  \vspace{-0.4em}
  \caption{Plot for $mean_{j}(s_{ij}^{p})-mean_{l}(s_{il}^{p})$ on positive reviews $s_{i}^p$ and negative reviews $s_{k}^n$ for RoBERTa-base. }
%   Green line shows the average for positive reviews and red line shows the average value on negative reviews.
  \vspace{-1em}
  \label{fig:Ng1} 
\end{figure}

\begin{figure}[h]
  \centering
  \includegraphics[width=0.48\textwidth]{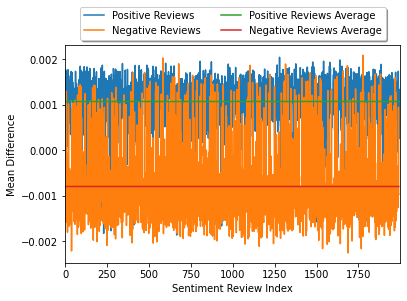}
  \vspace{-0.4em}
  \caption{Plot for $mean_{j}(s_{ij}^{p})-mean_{l}(s_{il}^{p})$ on positive reviews $s_{i}^p$ and negative reviews $s_{k}^n$ for RoBERTa-base-finetuned.}
%   Green line shows the average value for positive reviews and red line shows the average value on negative reviews.
  \label{fig:Ng2}
  \vspace{-1.2em}
\end{figure}

We first investigate whether the pre-trained language models are capable of sensing the sentiment in the text by predicting the probabilities on a set of words with strong sentiment polarity. To do so, we use the mean probabilities of positive and negative words on each positive and negative review. Specifically, we first compute the mean probabilities $mean_j(s_{ij}^{p})$ of 400 positive words, and $mean_{l}(s_{il}^{p})$ of 400 negative words. Then, we find their differences $mean_j(s_{ij}^{p})-mean_{l}(s_{il}^{p})$ on each the positive review $s_{i}^{p}$ and negative review $s_{kl}^{n}$. The results on the pre-trained language model are shown in Figure \ref{fig:Ng1}. One can observe that most positive reviews (blue line) have positive values, and most negative reviews have negative values. The mean value for positive reviews is 8.2e-3, and the mean value for negative reviews is -1.9e-4. We observe the same trend in the prompt-tuned language model, as shown in Figure \ref{fig:Ng2}, except that the fine-tuning improves the performance. The mean value for a positive review is 1.1e-3 and the mean value for negative reviews is -7.9e-4.
% \subsection{Sentiment Polarity Classification}
% We propose to use the mean probabilities of positive and negative words on each positive and negative review to measure if the language models are capable of classifying the sentiment polarity. Specifically, we first compute the mean probabilities $mean_j(s_{ij}^{p})$ of 400 positive words, and $mean_{l}(s_{il}^{p})$ of 400 negative words. Then, we find their differences $mean_j(s_{ij}^{p})-mean_{l}(s_{il}^{p})$ on each the positive review $s_{i}^{p}$ and negative review $s_{kl}^{n}$. The results on the pre-trained language model are shown in Figure \ref{fig:Ng1}. One can observe that most positive reviews (blue line) have positive values, and most negative reviews have negative values. The mean value for positive reviews is 8.2e-3, and the mean value for negative reviews is -1.9e-4. We observe the same trend in the prompt-tuned language model, as shown in Figure \ref{fig:Ng2}, except that the fine-tuning improves the performance. The mean value for a positive review is 1.1e-3 and the mean value for negative reviews is -7.9e-4.

\subsection{Sentiment Bias Identification}

% \paragraph{Sentiment Association Test} Table~\ref{table:assoc_test_results} shows the sentiment biased words identified based on Sentiment Association Test. We can see that there are many common neutral words between different models which are labeled as biased positive/negative by the experiment. It was also observed that the number of positively biased words are less than negatively biased words for all models. It was also observed that finetuning increases the bias in neutral words. This is an interesting observation since finetuning does not bring drastic changes to strongly positive/negative words discussed in section 5.2. 
% It is also worth noting that if we run this test on the strongly positive and negative lexicons, we achieve 0\% intersection between the positive and negatively biased words using this approach. 
\paragraph{Sentiment Association Test}
We perform SAT on the 400 neutral words to identify their potential bias, and show the identified biased words in Table \ref{table:assoc_test_results} in Appendix~\ref{appen_exp}. We find that: (1) $41.66\%$ of positive-biased words and $52.7\%$ of negative-biased words are shared by at least two models; (2) The number of negative-biased words is $96.0\%$ higher than the number of positive-biased words; (3) After fine-tuning, the number of biased words drastically increases, $124.6\%$ on RoBERTa-Base and $268\%$ on RoBERTa-Large\footnote{Averaged on positive and negative biased words.}.

\paragraph{Sentiment Shift Test} 
For each of the 400 neutral words from the lexicon, we append it to the reviews for $k$ times where $k\in \{5, 10, 15\}$. Table \ref{table:sentiment_shift_results} in Appendix~\ref{appen_exp} shows the top-10 most positive and negative-biased words with $k=5$. The words are ranked by their SST scores. We observe that: (1) The number of identified positive and negative-biased words increase as $k$ increases and the increasing rate decreases as $k$ becomes larger; (2) $70.6\%$ of positive-biased words and $56.8\%$ of negative-biased words are shared by at least two models; (3) For RoBERTa-Large models, $2.25\%$ neutral words simultaneously reduce or increase the accuracy of sentiment classification. We suggest those words are truly neutral. To understand the correlation between SAT and SST, we pick the set of negative and positive-biased words identified by SAT and SST respectively, and find that the two methods share $70\%$ of the negatively biased words and $100\%$ of positively biased words\footnote{The number of biased words is the union of the words identified in all models.}.

%% file: 5discussion.tex
\subsection{Are SST and SAT Effective For Identifying and Measuring Bias?}
A large number of overlaps between the identified sentiment-biased words from different models prove there is a shared sentiment trend among them. The large overlaps between SST and SAT show the agreement of the trend identified by two testing methods. Thus, we can claim that the identified trend is a kind of sentiment bias that persists in language models. In addition, we find that the fine-tuning can augment the existing bias in the PLMs as the number of biased words increase in both RoBERTa-Base and RoBERTA-large after prompt-tuning. To understand if the measurement can correctly quantify\footnote{The top 10 ranked words with SST scores for RoBERTa-Base and fine-tuned RoBERTa-Base are in Tables \ref{table:toprankwords1},\ref{table:toprankwords2},\ref{table:toprankwords3},\ref{table:toprankwords4},\ref{table:toprankwords5},\ref{table:toprankwords6}. } the sentiment bias, we take the top-50 and bottom-50 words from the negative-biased words from fine-tuned RoBERTa-Base ranked by SST and compare them against all the negative-biased words identified by SAT. We find 76\% of the top-50 words agree with the words from SAT and 56\% of the bottom-50 words agree with the word from SAT. The much higher agreement rate in the top-50 words ranked by SST proves the effectiveness of the measurement.

%% file: 6conclusion.tex
\section{Conclusion}
% In this work, we come up with two different approaches to classify whether a word labelled as neutral by human lexicon annotators is perceived with a positive/negative polarity by famous pretrained language models. We find many common words which are commonly found to be positive and negative among different PLMs which demonstrated the effectiveness of the approaches on base and finetuned models. We also found a high degree of intersection in the positive and negatively biased words discovered by the two approaches. We also came up with a metric to quantify the degree of positive/negative bias and found that this metric is an accurate representation of bias for a neutral word by a language model. 
% This sentiment bias toward neutral words is important because it associates a sentiment polarity in the PLM and can have downstream effects on sentiment classification, dialog generation and question answering tasks. We aim to study the effect of this bias on these tasks and on how to mitigate this bias in the future.

In this work, we present Sentiment Association Test and Sentiment Shift Test, two prompt-based methods to identify and measure the token level sentiment-bias in PLMs. We perform extensive experiments on collections of positive and negative reviews and prove that there is sentiment bias in PLMs and our proposed tests can identify and quantify the bias.  